\documentclass{article}
\usepackage{ijcai21}

\usepackage{times}
\usepackage{soul}
\usepackage{url}
\usepackage[hidelinks]{hyperref}
\usepackage[utf8]{inputenc}
\usepackage[small]{caption}
\usepackage{graphicx}
\usepackage{amsmath}
\usepackage{amsthm}
\usepackage{booktabs}
\usepackage{algorithm}
\usepackage{algorithmic}

\usepackage{helvet}
\usepackage{courier}
\usepackage{mdwmath}
\usepackage{mathrsfs}
\usepackage{multirow}
\usepackage{mdwtab}
\usepackage{amssymb}
\usepackage{array}
\usepackage{amsfonts}

\usepackage{eqparbox}
\usepackage{enumerate}
\usepackage{framed}
\usepackage{fancyhdr}
\usepackage{lipsum}
\usepackage{enumitem}

\title{Behavior Mimics Distribution: Combining Individual and Group \\Behaviors for Federated Learning}

\author{
Hua Huang$^1$
\and
Fanhua Shang$^{1,2}$\thanks{Corresponding authors.} \and
Yuanyuan Liu$^{1}$\And
Hongying Liu$^{1*}$
\affiliations
$^1$\! Key Lab of Intelligent Perception and Image Understanding of Ministry of Education, \\School of Artificial Intelligence, Xidian University, China\\$^2$\! Peng Cheng Lab, Shenzhen, China\\ \emails \{hhuang, fhshang, yyliu, hyliu\}@xidian.edu.cn
}

\begin{document}

\maketitle

\begin{abstract}
  Federated Learning (FL) has become an active and promising distributed machine learning paradigm. As a result of statistical heterogeneity, recent studies clearly show that the performance of popular FL methods (e.g., FedAvg) deteriorates dramatically due to the client drift caused by local updates. This paper proposes a novel Federated Learning algorithm (called \textbf{IGFL}), which leverages both \emph{\textbf{I}ndividual} and \emph{\textbf{G}roup} behaviors to mimic distribution, thereby improving the ability to deal with heterogeneity. Unlike existing FL methods, our IGFL can be applied to both client and server optimization. As a by-product, we propose a new \emph{attention-based} \emph{federated} \emph{learning} in the server optimization of IGFL. To the best of our knowledge, this is the first time to incorporate attention mechanisms into federated optimization. We conduct extensive experiments and show that IGFL can significantly improve the performance of existing federated learning methods. Especially when the distributions of data among individuals are diverse, IGFL can improve the classification accuracy by about 13\% compared with prior baselines.
\end{abstract}

\section{Introduction}
Federated learning is originally designed to allow massive mobile devices to collaboratively train a global statistical model. In recent years, it has emerged as a new paradigm for large-scale distributed machine learning without exchanging privacy data. The traditional distributed methods require that individuals upload their local data to a cloud parameter server (PS), and the server updates the model, where individuals can be remote devices or siloed data centers, such as banks and hospitals. An urgent problem is that most individuals are not willing to communicate their personal data, such as private photos or personal financial information, to an untrusted third party~\cite{mcmahan2017communication}.

General federated learning first distributes the aggregation model to each client by the parameter server. The clients update the model multiple times locally, and then communicate  the model updates  to the server. During the whole process, FL requires data not to be shared, which ensures the privacy of the clients. Statistical heterogeneity is a key and tough challenge for FL, which is usually not considered in distributed machine learning. Individuals with unique attributes make the collected data examples have different statistical distributions. Unfortunately, many recently proposed distributed machine learning methods are based on the independent and identically distributed (I.I.D.) assumption. This makes most of them directly applied to FL often ineffective~\cite{li2020federated}. In addition, similar to distributed machine learning, federated learning also suffers from expensive communication~\cite{konevcny2016federated}. With the development of modern mobile devices, the computational cost of the devices are usually far less than the communication cost~\cite{li2020federated}. A common communication-efficient solution is to use Local SGD~\cite{stich2019local,lin2019don}. This method allows clients to perform multiple local updates instead of once, and then communicate the results to the central server.

The FL algorithm based on Local SGD is the widely-used FedAvg~\cite{mcmahan2017communication}. However, the multiple local updates keep the client away from the global optimum. In extreme cases, each client reaches the local optimum, and then the server aggregates them. This is equivalent to ont-shot averaging, which does not work in non-IID setting~\cite{reddi2021adaptive}. Subsequently, researchers proposed a variety of new and improved algorithms. These methods can be roughly divided into two categories: i) \emph{client optimization}. The typical algorithm is SCAFFOLD~\cite{karimireddy2020scaffold}, which uses control variates at the stage of local updates, to alleviate the client drift. Inspired by variance reduction~\cite{NIPS2013_ac1dd209,shang:ssvrg}, \cite{liang2019variance} proposed a similar method. ii) \emph{server optimization}. These methods involve a server aggregation stage with momentum acceleration on the server as in FedAvgM~\cite{hsu2019measuring} or an adaptive optimizer~\cite{kingma2015adam} on the server as in FedAdam~\cite{reddi2021adaptive}. This category of algorithms regard the aggregation of the information from various individuals as a pseudo-gradient-based optimization process, which is called server optimization. In FedAvg, its server learning rate is always 1, which makes it lack of flexibility in server optimization.

Since the data is not shared, only the distributions of individuals which are different from the distribution of the whole group, are used for local updates. Server aggregation is actually a synchronization function that can eliminate the impact of this difference to a certain extent, depending on the frequency of communication. Therefore, this raises a question: \emph{Is there a better way to integrate individual distribution into group distribution instead of just relying on synchronization?} We answer this question in the \textbf{affirmative} and propose a new federated learning algorithm, called \emph{\textbf{I}ndividual} \emph{and} \emph{\textbf{G}roup} \emph{\textbf{F}ederated} \emph{\textbf{L}earning} (\textbf{IGFL}). IGFL leverages the behaviors (e.g., updates) of individuals and groups to mimic the corresponding distributions, and digests them in client optimization and server optimization.
In particular, inspired by the success of attention mechanisms~\cite{bahdanau2015neural}, we investigate the method of incorporating attention into federal learning and applied it to the server optimization of IGFL.

\subsection{Our Contributions}
We summarize our main contributions as follows.

$\bullet$ In order to reduce the adverse impacts of client drift caused by local updates, we propose a novel and unified federated learning algorithm (called \textbf{IGFL}). IGFL has an insight into the interplay between the behaviors and distributions of individuals and groups, that help to improve both client optimization and server optimization. IGFL with only our client or server optimization is called IGFL-C or IGFL-S.

$\bullet$  Moreover, taking the behaviors of individuals and groups into account for server optimization, and inspired by the success of attention mechanisms, we also propose a new \emph{\textbf{Attention-based}} \emph{\textbf{Federated}} \emph{\textbf{Learning}} (AFL) scheme. To the best of our knowledge, this is the first time to incorporate attention into federated optimization.

$\bullet$ Finally, We conduct extensive experiments on the CIFAR10 and EMNIST  data sets, and all the results show that the proposed algorithms exhibit superior performance to other algorithms in various situations. Especially, our algorithm can improve the classification accuracy by about 13\% compared with prior baselines (e.g., FedAvg) in highly heterogeneous (non-IID) settings.

\section{Related Work}
FedAvg~\cite{mcmahan2017communication} is a standard federated optimization algorithm, but it is tricky to deal with heterogeneous data. Recently, in order to better deal with non-IID data, researchers have proposed many improved methods. FedProx proposed in \cite{li2018federated} add a proximal term to the objective for stable updates.~\cite{zhao2018federated} presented a sharing strategy to improve performance, but it violated the basic requirement of federated learning. FedAvgM is a method of adding momentum on the server, proposed by \cite{hsu2019measuring}, and the authors also proposed a new way, which rely on Dirichlet distribution, to generate measurable federated data sets. SCAFFOLD~\cite{karimireddy2020scaffold} uses control variables to alleviate client drift, and it can be viewed as employing the idea of variance reduction on the client. Similar to SCAFFOLD, VRL-SGD~\cite{liang2019variance} also utilizes the method of variance reduction, but does not support client sampling. FedAdam~\cite{reddi2021adaptive} introduces an adaptive optimization method on the server side. Unlike the synchronous method, \cite{chen2020asynchronous} proposed an computationally efficient and asynchronous online method.

\section{Individual and Group Federated Learning}
For general federated learning problems, they can be expressed as the following problem:
\begin{equation}
\mathop{\min}_{w\in \mathbb{R}^d}f(w):=\frac{1}{P}\sum_{i=1}^{P}f_i(w),
\end{equation}
where $f_i(w):=\mathbb{E}_{x\sim \mathcal{P}_i}[f_i(w,x)]$ is the local loss function of the $i$-th client, $P$ is the total number of clients, and $\mathcal{P}_i$ denotes the distribution for the $i$-th local data. We assume $f_i$ is smooth. In this paper, we focus on the non-IID setting, where $\mathcal{P}_i$ may be very different (i.e., $\mathbb{E}_{\mathcal{P}_i}[f_i(w)]\neq f(w)$).

Our proposed \textbf{IGFL} algorithm is described in Algorithm \ref{alg:IGFL}. It includes two important schemes, \emph{IGFL-client} (Line 4) and \emph{IGFL-server} (Line 6), which are shown in Algorithms~\ref{alg:icfl-c} and~\ref{alg:icfl-s}, respectively. If we formulate the main update rules in Lines 4 and 6 into Eqs.\ (\ref{eq:local-updates}) and (\ref{eq:local-agg}), IGFL degenerates into FedAvg. We refer to the algorithms that only apply the client or server optimization scheme by IGFL-C (see Section~\ref{sec:igfl-c} for details) and IGFL-S (see Section~\ref{sec:afl} for details), respectively. The idea behind IGFL is to make full use of both individual and group behaviors to approximate distribution $\mathcal{P}$ in client optimization and server optimization.

\begin{algorithm}[t]
\caption{IGFL}
\textbf{Input}: $P$, $R$. \\
\textbf{Initialization}:$w_{ps}^0$, $\Delta w_k^0=0, k\in[P]$. \\
\textbf{Output}: $w_{ps}^{R}$.
\begin{algorithmic}[1] 
\FOR{$r=0,...,R-1$}
\STATE Sample subset $\mathcal{S}$ from clients;
\FOR{each client $i\in\mathcal{S}$ \textbf{in parallel}}
\STATE $\Delta w_i^{r+1}=$\textbf{IGFL-client}($w^{r}_{ps},\ i,\ \Delta w_i^r$);
\ENDFOR
\STATE $w_{ps}^{r+1}=$\textbf{IGFL-server}($\Delta w_i^{r+1}, \Delta w_i^{r}|i\in \mathcal{S}\}$);
\ENDFOR
\STATE \textbf{return} $w_{ps}^{R}$.
\end{algorithmic}
\label{alg:IGFL}
\end{algorithm}

\subsection{Review SGD and Local SGD}
To motivate the algorithms in this paper, we revisit the difference between SGD and Local SGD to illustrate the shortcomings of Local SGD. We assume that all clients are activated and the results obtained are easily extended to partly activation. For the $i$-th client in the $r$-th round, after pulling the global model (i.e., $w_i^r \leftarrow w_{ps}^r$), mini-batch SGD~\cite{shang:vrsgd} can be written in the following form:
\begin{equation}
w_i^{r+1}=w_i^{r}-\eta_l g_i(w_i^{r}),
\end{equation}
where $\eta_l$ is the client learning rate, $w_{ps}^r$ and $w_i^r$ are the parameters of the server and the $i$-th client in the $r$-th round, respectively. By aggregating $w_i^{r+1}$ on the server, we have
\begin{equation}
w_{ps}^{r+1}=\frac{1}{P}\sum_{i=1}^{P}w_i^{r+1}=w_{ps}^{r}- \frac{\eta_l}{P} \sum_{i=1}^{P} g_i(w_{ps}^{r}),
\end{equation}
where $P$ is the number of client. While Local SGD has the following form in the  $i$-th client:
\begin{equation}\label{eq:local-updates}
w_i^{r+1}=w_{ps}^r-\eta_l\sum_{t=0}^{T-1}g_i(w_{i}^{r,t}),
\end{equation}
where $T$ is the number of local steps. After aggregation,
\begin{equation} \label{eq:local-agg}
w_{ps}^{r+1}=w_{ps}^r-\frac{\eta_l}{P}\sum_{i=1}^{P}\sum_{t=0}^{T-1}g_i(w_{i}^{r,t}),
\end{equation}
where $w_{i}^{r,t}$ denotes the parameters of $i$-th client in the $r$-th round at $t$-th local step. Thanks to aggregation without delay, each update of SGD is based on the entire example set $x_{\sim \mathcal P}$. Unfortunately, the local updates of Local SGD depend only on the local example set $x_{\sim \mathcal P_i}$, resulting in the lack of interacting with other individuals in time. It tends to have a large deviation due to walking alone for a long time, especially as the number of local updates increases.

\begin{algorithm}[t]
\caption{IGFL-client}
\textbf{Input}: $w_{ps}^{r}$,\ $i$,\ $\Delta w_i^r$. \\
\textbf{Parameters}: $T$, $\eta_l$. \\
\textbf{Output}: $\Delta w_i^{r+1}$.
\begin{algorithmic}[1] 
\STATE $w_i^{r,0}=w_{ps}^{r}$; \;\;\;\;\;// pull global parameters
\STATE $\Delta w_{ps}=w_{ps}^r-w_{ps}^{r-1}$;
\FOR{$t=0,...,T-1$}
\STATE Compute gradient $g_i(w_i^{r,t})$;
\STATE $\Delta_{I}\!=\!-\eta_l g_i(w_i^{r,t})$, $\Delta_{G}\!=\!\frac{1}{|\mathcal{S}|}(\Delta_{I}\!-\!\frac{1}{T}\Delta w_i^r)\!+\!\frac{1}{T}\Delta w_{ps}$;\!
\STATE $w_i^{r,t+1}=w_i^{r,t}+\Delta_{I}+\Delta_{G}$;
\ENDFOR
\STATE $\Delta w_i^{r+1} = w_{i}^{r,T} - w_{ps}^r $;
\STATE \textbf{return} $\Delta w_i^{r+1}$.
\end{algorithmic}
\label{alg:icfl-c}
\end{algorithm}

\subsection{Use of Individual and Group Information}
\label{sec:igfl-c}
In order to make corrections, we want it to make full use of group information at every step forward, not just its own individual information. Thus, we have the following rules:
\begin{equation}
w_i^{r+1}\!=\!w_{ps}^r\!-\!\eta_l\sum_{t=0}^{T-1}g_i^{t*}, g_i^{t*}\!=\!\frac{1}{P}[g_i(w_{i}^{r,t}) + \sum_{k\ne i}^{P}g_k(w_{k}^{r,t})].\!\!
\label{eq:ideal-rule}
\end{equation}

However, as a result of the basic demands for privacy in FL, the individuals cannot get $\sum_{k\ne i}^{P}g_k(w_{k}^{r,t})$.
We define the approximation of individual and group behaviors as follows:
\begin{equation}
\Delta w_{ps}^r:=  w_{ps}^r-w_{ps}^{r-1},\;\;\; \Delta w_{i}^r:=w_i^{r}-w_i^{r-1},
\end{equation}
which are used to simulate the responses of individual distribution $\mathcal{P}_i$ and group distribution $\mathcal{P}$, respectively, and $\Delta w_{ps}^r\!=\!\frac{1}{P}\sum_{i=1}^{P}\Delta w_{i}^r$. We can assume that the gradient $g_i(w_i^{r,t})$ does not change quickly because $f_i(w)$ is smooth, and
$g_k(w_{i}^{r,t})\approx g_k(w_{i}^{r-1,t})$, similar to~\cite{karimireddy2020scaffold}. For the update rules (\ref{eq:local-updates}) and (\ref{eq:local-agg}), we have
\begin{equation} \label{eq:core}
\sum_{t=0}^{T-1}\sum_{k\ne i}^{P}g_k(w_{k}^{r,t})\approx \frac{P}{-\eta_l} \Delta w_{ps}^r - \frac{1}{-\eta_l}\Delta w_{i}^r.
\end{equation}

Now, plugging Eq.\ (\ref{eq:core}) back to  Eq.\ (\ref{eq:ideal-rule}), we get the gradient with respect to the approximate distribution of the group:
\begin{align}
  \hat{g}_i^t &= \frac{1}{P} \big[g_i(w_{i}^{r,t}) + \frac{P}{-\eta_lT}\Delta w_{ps}^r - \frac{1}{-\eta_lT}\Delta w_i^r\big] \notag\\
    &= \frac{1}{P}(g_i(w_{i}^{r,t})-\frac{1}{-\eta_lT}\Delta w_i^r)+ \frac{1}{-\eta_lT}\Delta w_{ps}^r.
\end{align}
Note that $\hat{g}_i^t$ represents the gradient that only depends on group behavior when updating locally. Furthermore, we hope that our algorithm not only depends on group behavior, but also ensures that individual specificity is not diluted. Therefore, in each step of the local update, clients also examine themselves sufficiently. Specifically, $g_i^t=\hat{g}_i^t+g_i(w_{i}^{r,t})$.  The detailed algorithm is given in Algorithm \ref{alg:icfl-c}.

\paragraph{Compared with FedAvgM.} If $\sum_{t=0}^{T-1}\sum_{k\ne i}^{P}g_k(w_{k}^{r,t})$ is added directly after $T$ local updates as additional distribution information, instead of spreading them to each local update, the update rule becomes:
\begin{equation} \label{moment}
w_i^{r+1}\leftarrow w_i^{r+1} + \Delta w_{ps}^r -\frac{1}{P} \Delta w_{i}^r.
\end{equation}
Obviously,  it is equivalent to injecting information directly on the server: $w_{ps}^{r+1}=\frac{1}{P}\sum_{i=1}^{P}w_{i}^{r+1}+ \frac{P-1}{P}\Delta w_{ps}^r$.  In terms of the rule, this is similar to FedAvgM proposed in \cite{hsu2019measuring}, which has the following form: $v\leftarrow \beta v + \Delta w_{ps}^{r+1}$, $w_{ps}^{r+1}\!=w_{ps}^r + v$. Note that the additional information injected by FedAvgM is not limited to the group behaviors of the previous round, but also those in the past. Besides, it will suffer from some major drawbacks if the information is injected at once: i) \emph{Volume},   $||\Delta w_{ps}^r||$ is much larger than the ordinary local update, which increases the uncertainty. ii) \emph{Timeliness},  more importantly, the internal gradient calculation still only contains individual behaviors. The timeliness of approximate group behaviors is further weakened. iii) \emph{Specificity}, the individual behaviors (i.e., $\Delta w_{j}^r$) are not utilized in this method.

\paragraph{Compared with SCAFFOLD.} Amazing and marvellous, $\hat{g}_i^t$ has a similar form as the update of SCAFFOLD~\cite{karimireddy2020scaffold}. Compared with SCAFFOLD, there is a subtle difference that IGFL-C changes the coefficient of $(g_i(w_{i}^{r,t})-\frac{1}{-\eta_lT}\Delta w_i^r)$ from 1 to ${1}/{P}$. Formally, ${1}/{P}$ comes from Eqs.\ (\ref{eq:core}) and (\ref{eq:ideal-rule}). As described in~\cite{defazio2014saga}, a variance reduction method is to use the estimator $\tilde{\nabla}_\xi:=\xi(X-Y)+\mathbb{E}Y$ to approximate $\mathbb{E}X$, where $\xi\in[0,1]$ and $Y$ is highly correlated with $X$. We have $\mathbb{E}\tilde{\nabla}_\xi=\xi \mathbb{E}X+(1-\xi)\mathbb{E}Y$,
Var$(\tilde{\nabla}_\xi)=\xi^2[$Var$(X)+$ Var$(Y)-2$ Cov$(X,Y)]$. Here, we set $g_i(w)\!=\!X$ and $\Delta w_i\!=\!Y$. SCAFFOLD uses $\xi\!=\!1$ to obtain an unbiased estimator. IGFL-C uses $\xi\!=\!1/P$ and has a non-zero bias. But the variance of IGFL-C is $1/P^2$ times the one of SCAFFOLD. In fact, due to the statistical heterogeneity, $\mathbb{E}_{\mathcal{P}_i}[g_i(w)]\neq \mathbb{E}_{\mathcal{P}}[g(w)]$, both of SCAFFOLD and IGFL-C are biased. Moreover, IGFL-C uses behaviors to simulate distributions, which can correct certain biases.

\section{Attention-based Federated Learning} \label{sec:afl}
The innovation of IGFL-client is mainly to apply the idea of simulating the responses of distributions by combining individual and group behaviors to the client optimization. We answer the following question in the \textbf{affirmative}: \emph{Whether this idea can also be applied to the server optimization to improve the ability of tackling statistical heterogeneity?} In this section, inspired by the attention mechanism, we introduce a novel \emph{\textbf{Attention-based}} \emph{\textbf{Federated}} \emph{\textbf{Learning}} (AFL) scheme as a tool for server optimization of IGFL. In essence, it measures the similarity between individual and group behaviors, or the similarity between two rounds of behaviors, and gives each client a new weight for summing them together. The attention mechanism has been widely used in many fields such as natural language processing~\cite{bahdanau2015neural} and computer vision~\cite{mnih2014recurrent}, and it is playing an increasingly important role. Next we will introduce how to use the attention mechanism innovatively to improve the capacity of federated learning to accommodate heterogeneity. To the best of our knowledge, this is the first time to apply attention mechanism to federated optimization.

As described in~\cite{vaswani2017attention}, the attention mechanism can be regarded as a mapping function about a set of key-value pairs and a query $Q$.  Specifically, the similarity between $Q$ and each $key$ is calculated to obtain a set of scores, which are used as weights. The output is a weighted sum of $values$. At the beginning of the $r$-th round of server optimization, the central server obtains the individual behaviors of selected clients $\{\Delta w_i^{r+1}|i\in \mathcal{S}\}$. In a general FL algorithm, the update rule of this round is the average of these behaviors: $w_{ps}^{r+1}\!=\!w_{ps}^r+\frac{1}{|\mathcal{S}|}\sum_{i\in \mathcal{S}}\Delta w_i^{r+1}$. In the proposed AFL, we treat these behaviors as a set of values $V$. By defining specific keys $K$ and queries $Q$, we can get the attention output $\Delta \hat {w}_i^{r+1}$. After that, the new parameters $w_{ps}^{r+1}$ are obtained by a mapping $\mathcal{M}(\{\Delta \hat{w}_i^{r+1}|i\in \mathcal{S}\}, w_{ps}^r)$, such as $w_{ps}^{r+1}=w_{ps}^r+\frac{1}{|\mathcal{S}|}\sum_{i\in \mathcal{S}}\Delta \hat{w}_i^{r+1}$. We compute $\Delta \hat{w}_i^{r+1}$ according to the following update rule:
\begin{equation}
\Delta \hat {w}_i^{r+1} = \sum_{j\in \mathcal{S}}\alpha_{ij}^r \Delta w_j^{r+1},\,
\alpha_{ij}^r= \frac{e^{\psi (Q_i^{r},\Delta w_{j}^{r+1})}}{\sum_{\tau}e^{\psi (Q_i^{r},\Delta w_{\tau}^{r+1})}},
\end{equation}
where $\psi(\cdot)$ is a similarity measurement function (e.g., we use dot product in this paper). Note that we set $K:=V=\{\Delta w_i^{r+1}|i\in \mathcal{S}\}$, which is consistent with the attention mechanism in other areas such as~\cite{bahdanau2015neural}. Regarding the selection of $Q$, we provide three different strategies, which are discussed in detail below.

\begin{algorithm}[t]
\caption{IGFL-server}
\label{alg:icfl-s}
\textbf{Input}: $\{\Delta w_i^{r+1}, \Delta w_i^{r} |i\in \mathcal{S}\}$. \\
\textbf{Output}: $w_{ps}^{r+1}$.
\begin{algorithmic}[1] 
\STATE $\Delta w_{ps}^{r+1}=\frac{1}{|\mathcal{S}|}\sum_{i\in\mathcal{S}}\Delta w_i^{r+1}$;
\FOR {each $i\in\mathcal{S}$}
\STATE $q_i\leftarrow \begin{cases}
                     \Delta w_i^{r+1}, & \mbox{Option I for self-attention} \\
                      \Delta w_{ps}^{r+1}, & \mbox{Option II for global-attention}\\
                      \Delta w_j^{r}, & \mbox{Option III for time-attention}
                    \end{cases}$
\STATE $\alpha_{ij}^r=\frac{e^{\psi(q_i, \Delta w_j^{r+1})}}{\sum_{\tau} e^{\psi(q_i, \Delta w_\tau^{r+1})}}$;
\STATE $\Delta \hat {w}_i^{r+1} = \sum_{j\in \mathcal{S}}\alpha_{ij}^r \Delta w_j^{r+1}$;
\ENDFOR
\STATE $w_{ps}^{r+1}=w_{ps}^r+\frac{1}{|\mathcal{S}|}\sum_{i\in \mathcal{S}}\Delta \hat{w}_i^{r+1}$;
\STATE \textbf{return} $w_{ps}^{r+1}$.
\end{algorithmic}
\end{algorithm}

\subsection{Self-Attention Federated Learning}
In this subsection, we introduce a new self-attention federated learning scheme, where $Q:=K=V=\{\Delta w_i^{r+1}|i\in \mathcal{S}\}$, similar to~\cite{vaswani2017attention}.
Hence,  the scores $\alpha_{ij}^r$ capture the similarity between individual $i$ and individual $j$ by measuring the similarity of their behaviors in this
round. This means that the individual that is more similar to individual $i$ should have more greater weight when representing individual $i$. The update rules can be formulated as follows:
\begin{equation*}
\Delta \hat{w}_i^{r+1} = \sum_{j\in \mathcal{S}}\alpha_{ij}^r \Delta w_j^{r+1},\,
\alpha_{ij}^r= \frac{e^{\psi (\Delta w_i^{r+1},\Delta w_{j}^{r+1})}}{\sum_{\tau}e^{\psi (\Delta w_i^{r+1},\Delta w_{\tau}^{r+1})}}.
\end{equation*}

\subsection{Global-Attention Federated Learning}
The attention score can be calculated not only from the similarity between individuals, but also from the similarity between individuals and groups. We propose a new global-attention federated learning scheme, i.e., $Q=\Delta w_{ps}^{r+1}$. Note that for any $i\in \mathcal{S}$, $Q_i$ does not change, i.e., there is only one query. As a result, for any $i\in \mathcal{S}$, $\Delta \hat{w}_i^{r+1}$ is equal. The update rules for server optimization become
\begin{equation*}
\!\!w_{ps}^{r+1}=w_{ps}^r+\sum_{i\in \mathcal{S}}\alpha_i^r \Delta w_i^{r+1},\; \alpha_{i}^r= \frac{e^{\psi (\Delta w_{ps}^{r+1},\Delta w_{i}^{r+1})}}{\sum_{\tau}e^{\psi (\Delta w_{ps}^{r+1},\Delta w_{\tau}^{r+1})}}.\!\!
\end{equation*}
The scores measure the similarity between individual and group behaviors. We should focus on an individual if its behavior is similar to the group behavior.

\subsection{Time-Attention Federated Learning}
The above two methods utilize the similarity between the individuals, or the individuals and the group in the current round. Time-attention FL sets $Q_i=\Delta w_j^{r-1}$ and provides a way to distribute attention by capturing the similarity between the behaviors of the current round and the previous round. One can see that the score $\alpha_{ij}^r$ is decoupled from $i$ and simplified to $\alpha_{j}^r$, similar to global-attention FL. Time-attention FL can be formulated as follows:
\begin{equation*}
\!\!w_{ps}^{r+1}=w_{ps}^r+\sum_{j\in \mathcal{S}}\alpha_j^r \Delta w_j^{r+1},\; \alpha_{j}^r= \frac{e^{\psi (\Delta w_{j}^{r},\,\Delta w_{j}^{r+1})}}{\sum_{\tau}e^{\psi (\Delta w_{j}^{r},\Delta w_{\tau}^{r+1})}}.\!\!
\end{equation*}
It can be interpreted that if the behaviors of an individual in two rounds are very similar, this means that the group behavior has little impact on individual correction, and this also implies that the individual is similar to the group.

\begin{table}
\vspace{-2mm}

\setlength{\tabcolsep}{16pt}
\centering
\begin{tabular}{lrr}
\toprule
Cross-silo  & $E=1$ & $E=5$ \\
\midrule
FedAvg       & 75.89  &  72.16     \\
FedAvgM            & 76.90  & 72.30       \\
SCAFFOLD    & 76.49  & 73.78      \\
FedAdam   & 76.29  & 71.22      \\
IGFL-C (ours)           & 77.25  & 74.69      \\
IGFL-S (ours)           & 76.17  & 73.61      \\
IGFL (ours)            & \textbf{78.08}   &\textbf{75.56}      \\
\bottomrule
\end{tabular}
\caption{Comparison of the average testing accuracies (\%) over the last 10\% rounds of each algorithm on CIFAR10 in the cross-silo setting after 5,000 or 3,000 communication rounds, which corresponds to $E\!=\!1$ and $E\!=\!5$, respectively. For IGFL-S and IGFL, we use the time-attention scheme to achieve the best performance.}
\label{tab:cross-silo-c}
\end{table}

\section{Experiments}
In this section, we demonstrate empirical evaluation of the proposed algorithms. We first verify the effectiveness
of our algorithm by training the model using convolutional networks on the CIFAR10~\cite{krizhevsky2009learning} and EMNIST~\cite{cohen2017emnist} data sets. Here we use the same network structure as in~\cite{mcmahan2017communication} and ~\cite{reddi2021adaptive}, respectively. We also conduct extensive experiments to further discuss IGFL. We implement all experiments in Ray~\cite{moritz2018ray} based Python, which is a flexible, high-performance distributed execution framework.

\subsection{Setup}
\paragraph{Data partitions.}
The non-IID populations are generated in two schemes: i) Sort-and-partition~\cite{mcmahan2017communication}. Each client has two shares with different labels, and each share is randomly selected from data partitions sorted by labels. ii) Dirichlet distribution~\cite{hsu2019measuring}. The training examples in each client are extracted by class following a categorical distribution generated by a Dirichlet distribution, $\boldsymbol c \sim Dir(\rho \boldsymbol q)$, where $\rho >0$ is a concentration parameter adjusting the heterogeneity among clients and $\boldsymbol q$ is a prior probability, assuming a uniform distribution.

\paragraph{Hyperparameter tuning and methods.}
For both CIFAR10 and EMNIST, we used the following parameters: batch size $B\!=\!100$, local epoch $E \!=\! \{1, 5\}$, and client selection rate $C \!=\! \{0.1, 1\}$, without step decay. We set the grid search range of client learning rate by $\eta_l \!\in\!\{10^{-3}, 3\times 10^{-3},...,10^{-1}, 3\times 10^{-1}\}$. We fixed the server learning rate to 1, except for FedAadam. We use the following methods as compared algorithms: FedAvg~\cite{mcmahan2017communication}, FedAvgM~\cite{hsu2019measuring}, SCAFFOLD~\cite{karimireddy2020scaffold}, and FedAdam~\cite{reddi2021adaptive}. Specifically, we set $\beta\!=\!0.9$ for FedAvgM, and adjust $\tau \in \{0.1, 0.01\}$ to achieve the best performance for FedAdam.

\subsection{Main Results}
We conduct extensive experiments to evaluate the performance of the proposed algorithms in the following two settings: i) \emph{Cross-silo setting}, in which we set $P \!=\!10$, the selection rate $C\!=\!1$, and adopt the sort-and-partition scheme. ii) \emph{Cross-device}, in which we set $P \!=\! 100$, $C \!=\! 0.1$, and use the Dirichlet distribution scheme.

\begin{table}
\vspace{-2mm}
\centering
\setlength{\tabcolsep}{6.19pt}
\begin{tabular}{llccc}
\toprule
&Cross-device & $\rho=1000$ & $\rho=1$ &$\rho=0.1$ \\
\midrule
\multirow{7}*{$E=1$} &FedAvg  &82.03   & 75.22 & 68.62    \\
    &FedAvgM     &82.21   & 77.34 & 70.12     \\
    &SCAFFOLD   &80.99   & 75.97 & --    \\
    &FedAdam   & 82.55  & 76.21 & 72.21   \\
    &IGFL-C (ours)        &83.56   &79.51  & 78.71    \\
    &IGFL-S (ours)        & 83.17  &79.14  & 73.09  \\
    &IGFL (ours)   & \textbf{84.01}   &\textbf{81.33}  & \textbf{79.45}    \\
\midrule
\multirow{7}*{$E=5$} &FedAvg  & 82.21  & 73.29 & 63.67    \\
    &FedAvgM     & 82.75  & 73.62 & 67.10     \\
    &SCAFFOLD   & 81.83  & 71.84 & --    \\
    &FedAdam   & 82.82  & 75.50 &  67.07   \\
    &IGFL-C (ours)          & 83.10  & 78.50 & 72.31    \\
    &IGFL-S (ours)        & 82.51  & 73.63 &  66.79  \\
    &IGFL (ours)   &  \textbf{84.13}  &  \textbf{78.63} &  \textbf{76.38}    \\
\bottomrule
\end{tabular}
\caption{Comparison of the average testing accuracy (\%) over the last 10\% rounds of each algorithm on CIFAR10 in the cross-device setting after 10,000 or 4,000 communication rounds, which corresponds to $E=1$ and $E=5$, respectively. For IGFL-S and IGFL, we use the global-attention scheme.}
\label{tab:cross-device}
\end{table}

\begin{figure}
\centering
\includegraphics[width=0.495\columnwidth]{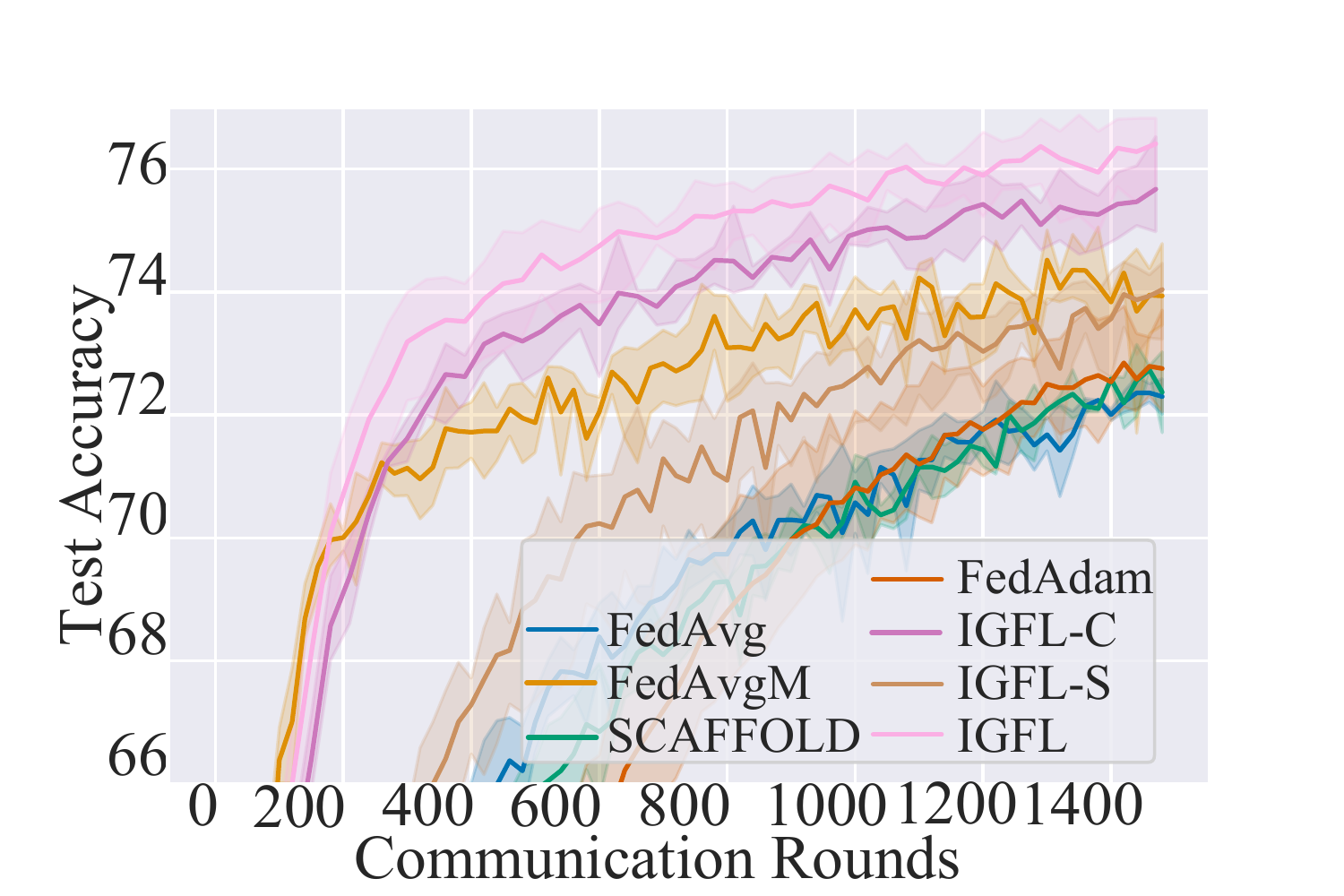}
\includegraphics[width=0.495\columnwidth]{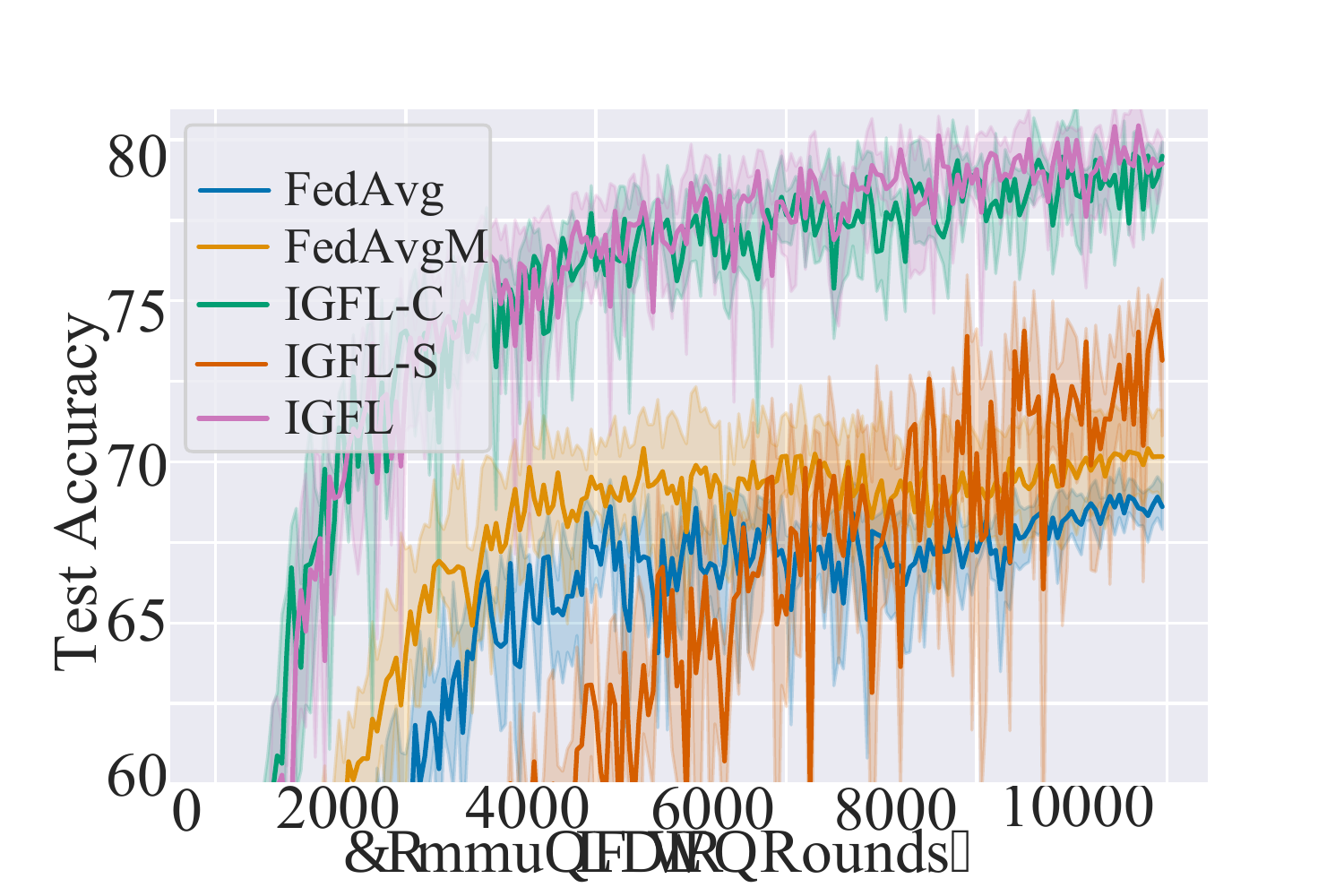}
\vspace{-5mm}

\caption {Test accuracies vs.\ the number of rounds of all the algorithms on CIFAR10 in the cross-silo (left) and cross-device (right) settings (here $E\!=\!1$ and $\rho\!=\!0.1$, best viewed in color).}
\label{fig:fig1}
\end{figure}

For the cross-silo setting, Table \ref{tab:cross-silo-c} shows the performance comparison of all the algorithms on CIFAR10 for visual classification tasks, when $E=1$ or $E=5$. Meanwhile, we give the test accuracy curve along with the number of rounds, as shown in Figure~\ref{fig:fig1}. Experimental results show that the proposed algorithms exhibit superior performance to other algorithms. Note that when encountering a larger number of local epochs, (i.e., $E=5$), the performance of FedAvgM degenerates and is only slightly better than that of FedAvg, while IGFL still has a significantly improvement over FedAvg. We will discuss this phenomenon in detail in Section~\ref{sec:avgm}.

In the cross-device setting, Tables ~\ref{tab:cross-device} and ~\ref{tab:cross-device-emnist} show the corresponding experimental results on CIFAR10 and EMNIST, respectively. The proposed methods achieve impressive improvements in terms of accuracy over other methods. In particular, it is exciting to see our algorithm improves accuracy by nearly \textbf{13\%} over the baseline method, (i.e., FedAvg), on CIFAR10 for the highly skewed non-IID case (e.g., $\rho=0.1$), while the improvement of other federated optimization algorithms (e.g., FedAvgM, FedAdam) is only about \textbf{4\%}.

\begin{table}
\vspace{-2mm}
\setlength{\tabcolsep}{20pt}
\centering
\begin{tabular}{lrr}
\toprule
Cross-device    & $\rho=1$ & $\rho=0.1$ \\
\midrule
FedAvg         & 77.72  & 70.83    \\
FedAvgM              & 78.07 & 75.37     \\
SCAFFOLD     & 78.08  & 68.20  \\
FedAdam     &  77.90  & 71.17 \\
IGFL-C (ours)             & 78.96 & 72.61  \\
IGFL-S (ours)           & 78.88 & 74.63  \\
IGFL (ours)         &  \textbf{81.17}  & \textbf{77.27} \\
\bottomrule
\end{tabular}
\vspace{-2mm}

\caption{Comparison of the average testing accuracies (\%) over the last 10\% rounds of each algorithm on EMNIST in the cross-device setting after 1,500 communication rounds. For IGFL-S and IGFL, we use the self-attention scheme.}
\label{tab:cross-device-emnist}
\end{table}

\subsection{Effectiveness of Amortization Against FedAvgM} \label{sec:avgm}
We design and run experiments to discuss the relationship between IGFL-C and FedAvgM to illustrate that IGFL-C can significantly reduce client drift by spreading the approximate group distribution information to each local update, rather than at the end of each round. We change $T$, which depends on $B$ and $E$. Specifically, we set $B=\{20, 100\}$, $E=\{1, 5\}$. The experimental results are shown in Table \ref{tab:fedavgm-ig}. It can be found that as the number of local updates increases, the performance difference between them becomes more and more obvious, which shows the advantages of the amortization strategy used in IGFL-C when encountering a large local epoch number.

\begin{table}
\centering
\vspace{-2mm}
\setlength{\tabcolsep}{8.5pt}
\begin{tabular}{lccc}
\toprule
$T$ ($B$, $E$)  & FedAvgM & IGFL-C & $\delta$ \\
\midrule
50 ($B\!=\!100$, $E\!=\!1$)       & 76.90  &  77.25 & 0.35     \\
250 ($B\!=\!20$, $E\!=\!1$)       & 71.83 & 73.35 &1.52         \\
250 ($B\!=\!100$, $E\!=\!5$)     & 72.30  & 74.69 & 2.39      \\
1250 ($B\!=\!20$, $E\!=\!5$)   & 68.54  & 71.54   & 3.00   \\
\bottomrule
\end{tabular}
\vspace{-2mm}

\caption{Comparison of the testing accuracies (\%) of FedAvgM and IGFL-C as varying $T$ in the cross-silo setting.}
\label{tab:fedavgm-ig}
\end{table}

\subsection{Comparison of Attention-based Methods}
We first compare the performance of the three proposed attention schemes, as shown in Table \ref{tab:attention}. The experimental results suggest that specific settings may require specific attention to achieve the best performance. For example, in the case of no client sampling, time-attention performs very well, because under the sampling conditions, the previous behaviors saved by different clients correspond to different rounds. Global-attention has excellent performance on CIFAR10 (10 classes), but not well on EMNIST (62 classes), which implies that its performance may be affected by the number of classes. More experiments are needed for conclusions.

\begin{figure}[t]
\centering
\includegraphics[width=0.489\columnwidth]{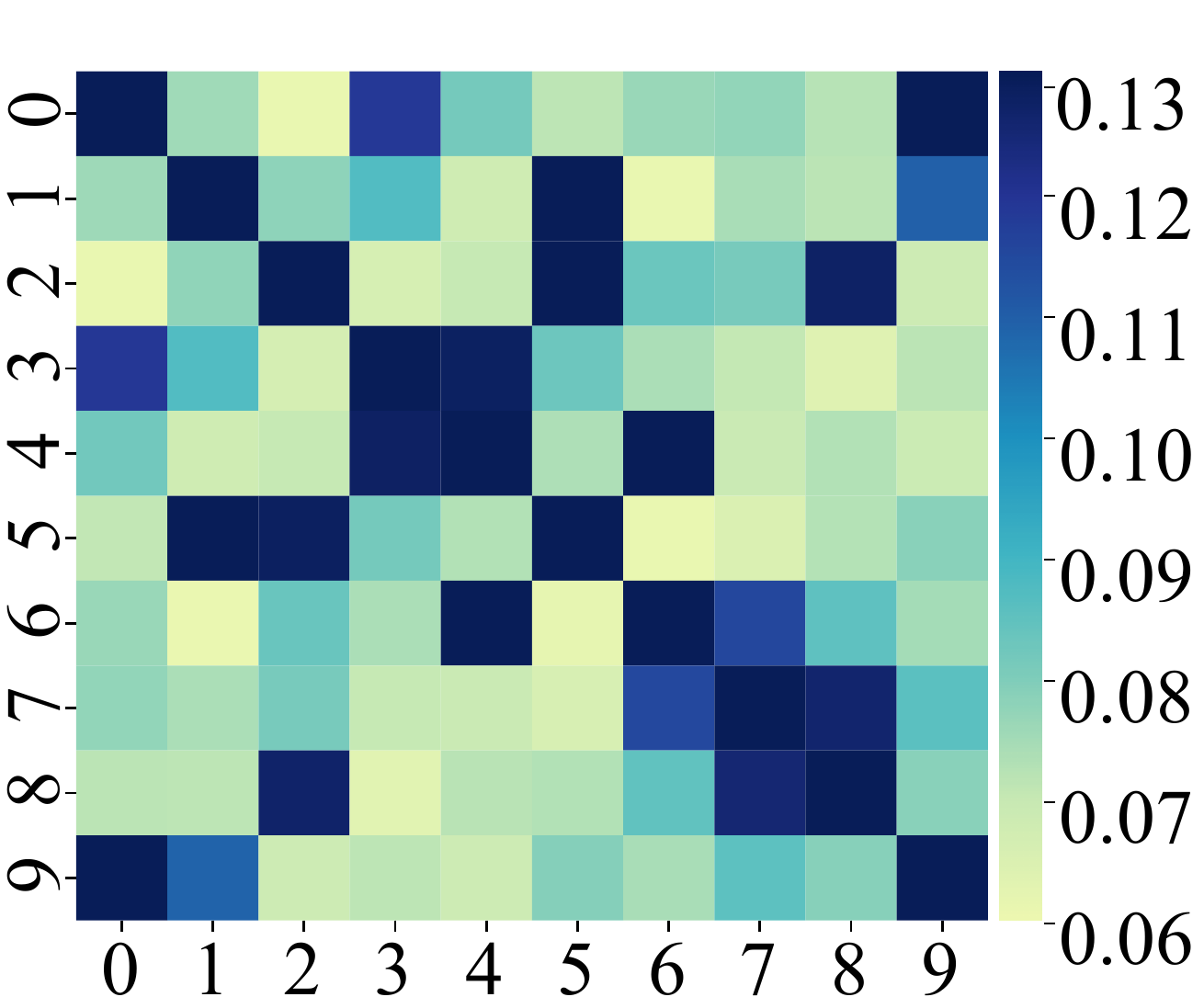}
\includegraphics[width=0.489\columnwidth]{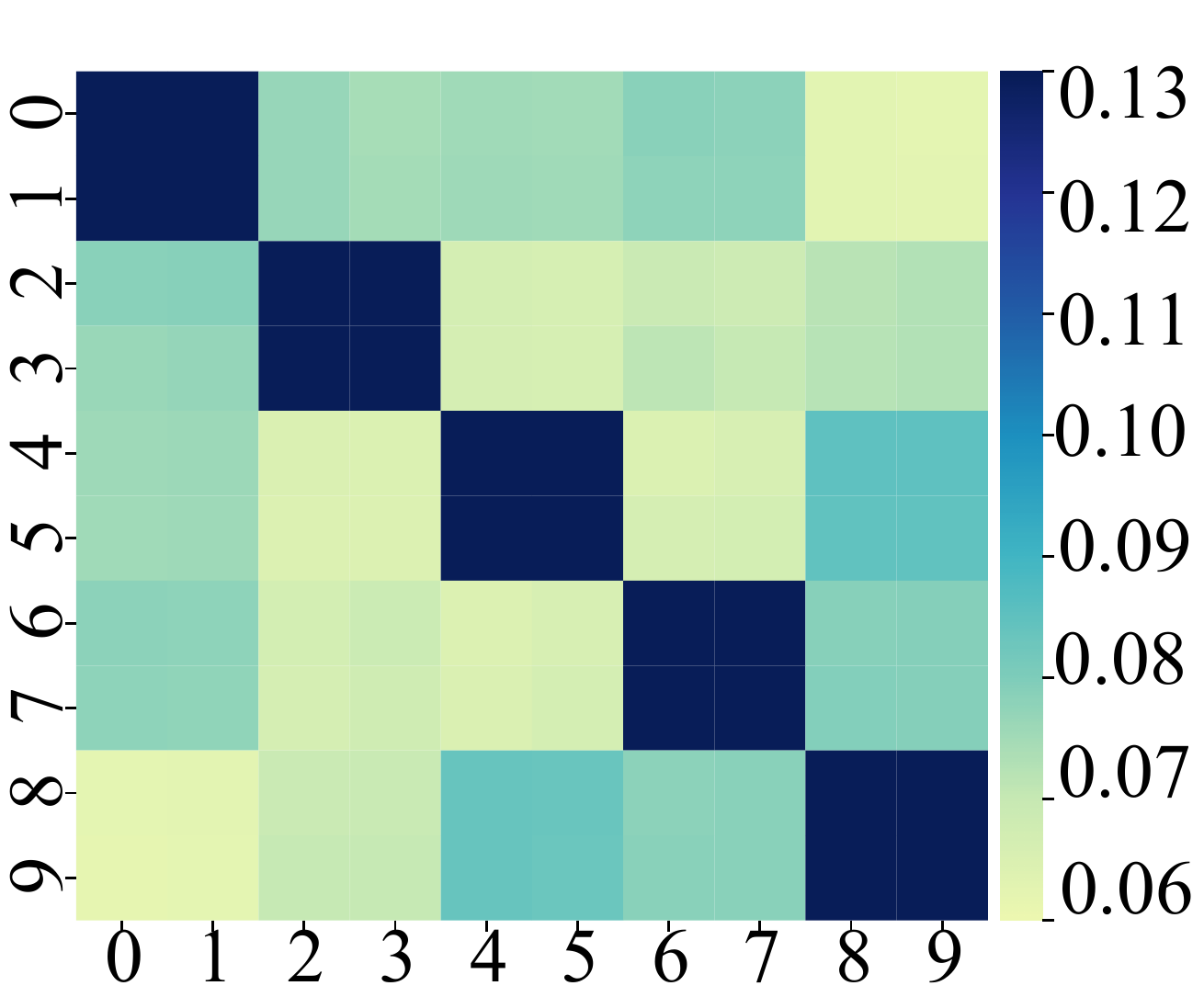}
\vspace{-2mm}

\caption{Visualization of how self-attention works, where the attention scores are the average of all (1,000 rounds) scores. The populations corresponding to random example (left) and paired example (right) are $\{$(0,9),(1,5),(2,8),(3,0),(4,3),(5,2),(6,4),(7,6),(8,7),(9,1)$\}$ and $\{$(0,1),(0,1),(2,3),(2,3),...,(8,9),(8,9)$\}$, respectively, where $(i,j)$ denotes the client assigned the $i$-th and $j$-th labels.}
\label{fig:attention}
\end{figure}

Moreover, the visualization of how the behaviors mimic the distributions is given in Figure \ref{fig:attention}. We set two specific populations for CIFAR10: i) random population by using the standard sort-and-partition, and ii) paired population, in which every two clients have the same label distribution. We generate 50 different random populations for experiments, and calculate the matching rate between client pairs with the same label and darker coordinate pairs on the heat map, which is up to 96\%. Experimental results show that self-attention can indeed capture the similarity of distribution between clients, by measuring the similarity of individual behaviors.

\begin{table}
\centering
\begin{tabular}{lccccc}
\toprule
\!Setting & $E$ &$\rho$ & GA & SA & TA \\
\midrule
\multirow{2}*{\!CIFAR10 silo} & $E\!=\!1$&--       & 77.48  &  76.46 & \textbf{78.08}     \\
            &$E\!=\!5$ &-- & 74.50 & 73.91 & \textbf{75.56}         \\
\midrule
\multirow{6}*{\!CIFAR10 device\!\!}&\multirow{3}*{$E\!=\!1$}&1000     & 84.01  & 83.30 & \textbf{84.17}      \\
            &       &1     & \textbf{81.33}  & 80.07 & 79.95      \\
            &       &0.1     & \textbf{79.45}  & 79.03 & 76.83      \\
            &\multirow{3}*{$E\!=\!5$}  &1000  & \textbf{84.13}  & 83.16   &  80.06  \\
            &       &1     & \textbf{78.63}  & 78.56 & 77.94      \\
            &       &0.1     & \textbf{76.38}  & 75.92 & 74.89     \\
\midrule
\multirow{2}*{\!EMNIST device\!\!}&\multirow{2}*{$E\!=\!1$}&1     & 78.51  & \textbf{81.17} & 78.73     \\
            &       &0.1     & 70.45  & \textbf{77.27} & 72.57      \\
\bottomrule
\end{tabular}
\vspace{-2mm}

\caption{Testing accuracies (\%) of our three attention schemes in different settings. Here, GA, SA and TA denote the global-attention, self-attention, and time-attention schemes, respectively.}
\label{tab:attention}
\end{table}

\section{Conclusions}
In this paper, we proposed a novel federated learning algorithm (called IGFL) with both a new client optimizer and a new server optimizer  to alleviate the statistical heterogeneity for federated learning. It leverages individual behavior as an estimate of the data distribution of individual, and complements the distribution of groups that are invisible but crucial during local updates, thus reducing the impact of client drift on FL algorithms. Different from other solutions, the technique that behavior mimic distribution can be used in both client optimization and server optimization. As a by-product, we also presented two federated learning algorithms with the proposed client optimizer or server optimizer, which are called IGFL-C and IGFL-S, respectively. Moreover, for our server optimizer, we proposed a federated learning optimization scheme (called AFL) based on the proposed attention mechanisms. The principle behind AFL is that attention captures the similarity of individual distributions. We performed extensive experiments to demonstrate that the proposed algorithms have significant improvements over other recently proposed algorithms, especially in the highly skewed non-IID case.

\section*{Acknowledgments}
This work was supported by the National Natural Science Foundation of China (Nos.\ 61876220, 61876221, 61976164, 61836009 and U1701267), the Program for Cheung Kong Scholars and Innovative Research Team in University (No.\ IRT\_15R53), the Fund for Foreign Scholars in University Research and Teaching Programs (the 111 Project) (No.\ B07048), and the National Science Basic Research Plan in Shaanxi Province of China (No.\ 2020JM-194).

\bibliographystyle{named}
\bibliography{ijcai21}

\end{document}